%% file: main.tex
\documentclass[runningheads]{llncs}
\input{preamble.tex}
\begin{document}
\title{Spatial Autoregressive Modeling of DINOv3 Embeddings for Unsupervised Anomaly Detection}
\titlerunning{Spatial Autoregressive Modeling of DINOv3 Embeddings for UAD}

\maketitle              
\begin{abstract}
DINO models provide rich patch-level representations that have recently enabled strong performance in unsupervised anomaly detection (UAD). 
Most existing methods extract patch embeddings from ``normal'' images and model them independently, ignoring spatial and neighborhood relationships between patches. 
This implicitly assumes that self-attention and positional encodings sufficiently encode contextual information within each patch embedding. 
In addition, the normative distribution is often modeled as memory banks or prototype-based representations, which require storing large numbers of features and performing costly comparisons at inference time, leading to substantial memory and computational overhead.
In this work, we explicitly model spatial and contextual dependencies between patch embeddings using a 2D autoregressive (AR) model, allowing us to use a simple and efficient model for normative modeling. 
Instead of storing embeddings or clustering prototypes, our approach learns a compact parametric model of the normative distribution via an AR convolutional neural network (CNN). 
At test time, anomaly detection reduces to a single forward pass through the network and enables fast and memory-efficient inference.
We evaluate our method on the BMAD benchmark, comprising three medical imaging datasets, and the VisA dataset, comprising 12 industrial object categories, and compare it against existing methods, including recent DINO-based approaches.
Experimental results demonstrate that explicitly modeling spatial dependencies achieves competitive anomaly detection performance while substantially reducing inference time and memory requirements. 
Code is available at the project page: \url{https://eerdil.github.io/spatial-ar-dinov3-uad/}.


\end{abstract}

\section{Introduction}
Anomaly detection (AD) in images aims to identify pixels or regions that deviate from normal patterns, such as lesions in medical imaging or structural defects in industrial inspection. Supervised approaches require both normal and abnormal images with pixel-level annotations, which demand substantial manual effort and domain expertise—particularly in medical imaging. Moreover, supervised training faces challenges such as class imbalance and the underrepresentation of rare anomaly types, potentially leading to biased models.

To address these limitations, unsupervised anomaly detection (UAD) methods learn normal anatomical patterns solely from healthy samples, without requiring anomalous data during training. By modeling the distribution of healthy anatomy, UAD methods detect deviations from the learned normative distribution at test time. Beyond detection accuracy, computational efficiency is crucial for real-world deployment, especially in clinical settings where time and hardware resources are limited. Therefore, memory-efficient and fast methods are highly desirable. In particular, methods that avoid storing large sets of training embeddings or performing costly nearest-neighbor search at inference time are attractive for scalable deployment.

Early UAD approaches are reconstruction-based, training models to reconstruct normal anatomical structures and using reconstruction residuals as anomaly scores. Autoencoders and VAEs \cite{chen2018unsupervised,zimmerer2019unsupervised} model normal data in a latent space, while GAN-based \cite{schlegl2019f} and diffusion-based methods \cite{wyatt2022anoddpm} aim to improve reconstruction fidelity. Although conceptually simple, reconstruction-based approaches may generalize too well and partially reconstruct anomalous regions.

Instead of reconstructing images, feature-based methods model embeddings extracted from pretrained networks. Knowledge distillation-based frameworks \cite{wang2021student,deng2022anomaly} detect discrepancies between teacher and student features. Other approaches estimate feature distributions using Gaussian models \cite{defard2021padim}, normalizing flows \cite{gudovskiy2022cflow}, clustering \cite{lee2022cfa}, or memory banks~\cite{cohen2020sub,roth2022towards,huang2024prototype}. Lastly, there exist methods that model the latent space of vector-quantized VAEs with autoregressive models~\cite{marimont2021anomaly,pinaya2022unsupervised}

Recent foundation models such as DINO \cite{oquab2023dinov2,simeoni2025dinov3} provide rich patch-level representations. DINO embeddings capture global context through self-attention and have proven highly transferable to anomaly detection~\cite{damm2025anomalydino,schulthess2025anomaly}. AnomalyDINO \cite{damm2025anomalydino} exploits these representations by storing embeddings of healthy patches in a memory bank and performing nearest-neighbor search at test time. While effective, this strategy can be memory-intensive, since a large number of normal patch embeddings must be retained, and inference requires costly feature matching. DINO-DPMM \cite{schulthess2025anomaly} addresses this limitation by modeling the distribution of healthy DINO embeddings using a Dirichlet process mixture model (DPMM). Instead of storing all training embeddings, their method represents the normal feature distribution through a set of mixture components or cluster centers. This reduces the memory burden compared with full memory-bank storage. 
However, the DINO-DPMM models the normal feature distribution less precisely than the memory bank in AnomalyDINO, sacrificing a bit of accuracy for substantial runtime gains.
Moreover, both AnomalyDINO and DINO-DPMM treat DINO patch embeddings primarily as individual feature samples. They estimate marginal feature distributions over healthy patches while largely disregarding the structured two-dimensional organization of the embedding grid. Although DINO self-attention ensures that each patch embedding is influenced by global image context, this does not by itself amount to explicitly modeling the joint spatial distribution of features across the patch grid.

Dinomaly \cite{guo2025dinomaly} follows a feature reconstruction/distillation paradigm, using a pretrained DINOv2 Vision Transformer as an encoder and training a Transformer-based decoder to reconstruct intermediate DINO features. This yields a compact learned model of healthy feature embeddings and avoids explicit memory-bank search. Through its attention-based decoder, Dinomaly can also exploit relationships among patch tokens. However, its normality model is still based on feature reconstruction: spatial dependencies are learned implicitly through the decoder architecture rather than explicitly parameterized as a conditional density over the two-dimensional DINO feature grid. In addition, Transformer-based decoders are generally heavier than lightweight convolutional alternatives, which can be limiting in resource-constrained deployment settings.

In this work, we propose a simple lightweight CNN-based autoregressive (AR) framework \cite{xiong2024autoregressive,sun2024autoregressive} that preserves the 2D grid structure of DINO embeddings and explicitly models conditional dependencies between patches. By capturing joint spatial interactions in feature space, our approach provides a compact and expressive model of normal anatomy while remaining computationally efficient. Unlike memory-bank methods, inference requires only a single forward pass, eliminating the need for costly nearest-neighbor search. Unlike Transformer-based reconstruction frameworks, our CNN-based AR model offers a lightweight way to exploit spatial structure in DINO feature grids.

Previous work has shown that likelihoods assigned by autoregressive models in pixel space can be unreliable for out-of-distribution detection, as they may be dominated by low-level image statistics rather than high-level semantic content~\cite{nalisnick2018do,ren2019likelihood,serra2019input,schirrmeister2020understanding}. In contrast to directly modeling image intensities, we propose to model DINO features, which encode richer semantic information and are therefore expected to reduce the sensitivity of likelihood estimates to such low-level statistics.

AR modeling of patch embeddings raises the question of how much spatial context should be incorporated when predicting each patch. In anatomically structured data, neighboring embeddings may exhibit strong short-range redundancy, allowing the model to rely primarily on local interpolation, similar to the observation in~\cite{umapathi2022shaken}. To study the influence of spatial scale in this setting, we incorporate dilated convolutions as a simple architectural modification that enlarges the receptive field without increasing computational cost. This design enables the model to capture broader contextual dependencies while preserving the autoregressive formulation, and allows us to investigate how different spatial modeling scales affect anomaly detection performance.

\section{Method}
\subsection{Autoregressive Modeling of DINO Features}
\begin{figure*}[t]
    \centering
    \includegraphics[width=0.9\linewidth]{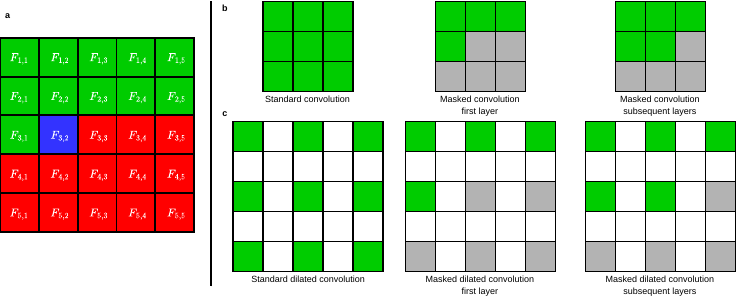}
    \caption{Illustration of \textbf{(a)} raster-scan autoregressive factorization over the embedding grid where the \textcolor{blue}{blue} cell indicates the current prediction target, \textcolor{green}{green} cells denote preceding (conditioned) embeddings, and \textcolor{red}{red} cells correspond to future spatial locations which are treated as unobserved to predict the \textcolor{blue}{blue} embedding, \textbf{(b)} Standard and masked convolutional kernels without dilation, and \textbf{(c)} corresponding dilated convolutional kernels. Grey cells indicate masked weights that prevent access to future spatial positions.}
    \label{fig:raster_scan_and_conv}
\end{figure*}
Let $x \in \mathbb{R}^{H \times W}$ denote an input image drawn from the distribution of healthy samples $p_\text{data}(x)$. A pretrained DINO vision transformer $\Phi$ extracts patch-level embeddings $F = \Phi(x) \in \mathbb{R}^{H_p \times W_p \times D}$, arranged on a 2D grid of size $H_p \times W_p$, where $D$ is the embedding dimension and $F_{i,j} \in \mathbb{R}^D$ denotes the embedding at location $(i,j)$.

We model the joint distribution over the embedding grid using an AR factorization \cite{tomczak2024autoregressive}:
\begin{equation}
    p(F) = \prod_{i,j} p(F_{i,j} \mid F_{<i,j}),
\label{eq:ar_factorization}
\end{equation}
where $F_{<i,j}$ denotes embeddings preceding $(i,j)$ under a raster-scan ordering (row-major traversal from top-left to bottom-right) as shown in Fig.~\ref{fig:raster_scan_and_conv}(a). This factorization enables structured density modeling over the two-dimensional embedding grid by capturing conditional dependencies across spatial locations.

Since prior work found normalized DINO features to be more expressive than their unnormalized counterpart~\cite{schulthess2025anomaly}, we model each conditional distribution using von Mises--Fisher (vMF) distribution:
\begin{equation}
p(F_{i,j} \mid F_{<i,j}) = \text{vMF}(F_{i,j} \mid \mu_{i,j}, \kappa),
\end{equation}
where $\kappa > 0$ is a concentration parameter and $\mu_{i,j}$ is predicted by a neural network $f_\theta(F)$, where the architecture enforces that $\mu_{i,j}$ depends only on $F_{<i,j}$. 

The model is trained on healthy samples by minimizing the negative log-likelihood:
\begin{equation}
\mathcal{L}(\theta)
=
- \frac{1}{|\mathcal{X}_\text{train}|}
\sum_{x \in \mathcal{X}_\text{train}}
\sum_{i,j}
\log p(F_{i,j} \mid F_{<i,j}; \theta),
\end{equation}
which is equivalent to minimizing cosine loss with the vMF assumption.
At test time, anomaly scores for each patch are computed from the corresponding conditional negative log-likelihood values:
\begin{equation}
A_{i,j} = -\log p(F_{i,j} \mid F_{<i,j}).
\end{equation}
Note that $\kappa$ acts as a constant shift and scale factor on $A_{i,j}$, however since the relative ordering of the anomaly scores is preserved, the choice of $\kappa$ is arbitrary for anomaly scoring.
All patch-level anomaly scores are obtained in a single forward pass through $f_\theta$. The pixel-level anomaly map is created by bilinear interpolation of the patch-level anomaly map. The architectural design enabling parallel AR evaluation is described in the following section.

\subsection{CNN-based Autoregressive Architecture}
\label{subsec:cnn_arch}
The AR factorization in Eq.~\eqref{eq:ar_factorization} requires that the prediction at location $(i,j)$ depends only on the preceding embeddings $F_{<i,j}$. 
Instead of performing sequential evaluation, we enforce this AR constraint using masked convolutions, enabling parallel training and inference similar to PixelCNN-style models \cite{van2016pixel}.

Given the embedding grid $F \in \mathbb{R}^{H_p \times W_p \times D}$, the network $f_\theta$ is implemented as a stack of masked convolutional layers. 
For a convolutional kernel centered at $(i,j)$, weights corresponding to spatial locations below the current row and to the right within the same row are set to zero, ensuring that the prediction does not access future embeddings under the raster-scan ordering. 

In the first convolutional layer, we additionally mask the center element of the kernel, preventing direct access to the current embedding $F_{i,j}$. 
In subsequent layers, the center position is unmasked, since intermediate feature maps no longer contain direct access to the original input but instead consist of already masked representations. 
Fig.~\ref{fig:raster_scan_and_conv}(b) illustrates the masking patterns applied in the first and subsequent convolutional layers.
This design enforces the AR constraint while enabling efficient parallel computation over the entire embedding grid instead of computationally expensive sequential evaluation \cite{van2016pixel}.

A key challenge when AR modeling of DINO embeddings is that these features are already globally contextualized through self-attention. 
As a result, neighboring embeddings often exhibit strong local correlations. 
A CNN may therefore learn to predict each embedding primarily from its immediate spatial neighbors, effectively performing short-range interpolation, as observed in~\cite{umapathi2022shaken}. 
Such behavior can reduce anomaly sensitivity particularly for large anomalies, as structurally abnormal regions may still be predictable from surrounding normal context.

To mitigate this issue, we introduce dilated convolutions \cite{yu2015multi} in the AR CNN. 
Dilated convolutions expand the receptive field without increasing the number of parameters by introducing gaps between kernel elements. 
This increases the effective receptive field under the AR constraint, enabling the model to capture longer-range spatial dependencies in the embedding grid.
The combined masking and dilation strategy is illustrated in Fig.~\ref{fig:raster_scan_and_conv}(c).

\section{Experiments and Results}

\subsection{Experimental Setup}
\label{subsec:exp_setup}

\paragraph{Model Architecture:} We employ the small (S) variant of DINOv3 as a feature extractor $\Phi$, initialized with officially released pretrained weights\footnote{\url{https://github.com/facebookresearch/dinov3} under the Apache 2.0 License.}, resulting in 384-dimensional patch embeddings. 
The AR network $f_\theta$ consists of five masked convolutional layers with a dilation factor of 4.
The model is trained using the AdamW optimizer \cite{loshchilov2017decoupled} with a batch size of 64 and a learning rate $10^{-3}$. 
Following~\cite{damm2025anomalydino,schulthess2025anomaly}, input images are resized to $448 \times 448$ to obtain smoother anomaly maps in DINO-based methods.
Model selection is performed based on the highest log-likelihood on the normal validation set.

\paragraph{Datasets:} We evaluate our method on the anomaly segmentation datasets provided by the BMAD benchmark \cite{bao2024bmad}, which include 2D images of brain MRI from BraTS2021 \cite{baid2021rsna}, liver CT from BTCV \cite{landman2015miccai} and LiTs \cite{bilic2023liver}, and retinal OCT from RESC \cite{hu2019automated}. 
BraTS2021 contains 8{,}179 normal and 3{,}119 anomalous images. 
The combined liver CT datasets comprise 2{,}468 normal and 733 anomalous images. 
RESC includes 5{,}383 normal and 834 anomalous images.
A separate model is trained for each dataset using only normal samples for training and validation.
Additional experiments on VisA dataset \cite{zou2022spot} are presented in Appendix~\ref{sec:app_visa}.

\paragraph{Evaluation:} We compare our approach against state-of-the-art UAD methods using pixel-level Area Under the Receiver Operating Characteristic curve (AUROC) and Area Under the Precision–Recall curve (AUPR).
The results are averaged over three random seeds, except AnomalyDINO which is deterministic.
Following \cite{schulthess2025anomaly}, evaluation is performed at the original image resolution.
In addition to detection performance, we report computational efficiency in terms of inference time per 2D image and peak total (CPU $+$ GPU) memory consumption during inference. 
Runtime is measured with batch size 1 and averaged over the entire test set after a short warm-up phase. The runtime measurements include the feature extraction with the respective backbones.
All experiments are performed on an NVIDIA A6000 GPU (48 GB VRAM).

\paragraph{Baselines:} We compare our method with a diverse set of UAD approaches, including reconstruction-based methods (DRAEM \cite{zavrtanik2021draem}, UTRAD \cite{chen2022utrad}, Dinomaly~\cite{guo2025dinomaly}), student–teacher methods (RD4AD \cite{deng2022anomaly}, STFPM \cite{wang2021student}), feature-distribution modeling approaches (PaDiM \cite{defard2021padim}, CFLOW \cite{gudovskiy2022cflow}, CFA \cite{lee2022cfa}), and memory-bank-based methods (PatchCore \cite{roth2022towards}, AnomalyDINO \cite{damm2025anomalydino}, DINO-DPMM \cite{schulthess2025anomaly}). 
We used the WideResNet50~\cite{Zagoruyko2016Wide} backbone for STFPM, PaDiM and PatchCore.
Results for PatchCore on BraTS2021 are not available since this experiment requires more than 80~GB of VRAM.
For AnomalyDINO, we report the full-shot setting.
Dinomaly~\cite{guo2025dinomaly}, AnomalyDINO~\cite{damm2025anomalydino} and DINO-DPMM~\cite{schulthess2025anomaly} were originally evaluated using DINOv2-S embeddings. To ensure a fair comparison, we report their published results with DINOv2-S as well as DINOv3-S embeddings.
Baseline results are obtained using official implementations and the BMAD benchmark implementations \cite{bao2024bmad}.

\subsection{Results}
\begin{figure}[t]
\centering
\includegraphics[width=\linewidth]{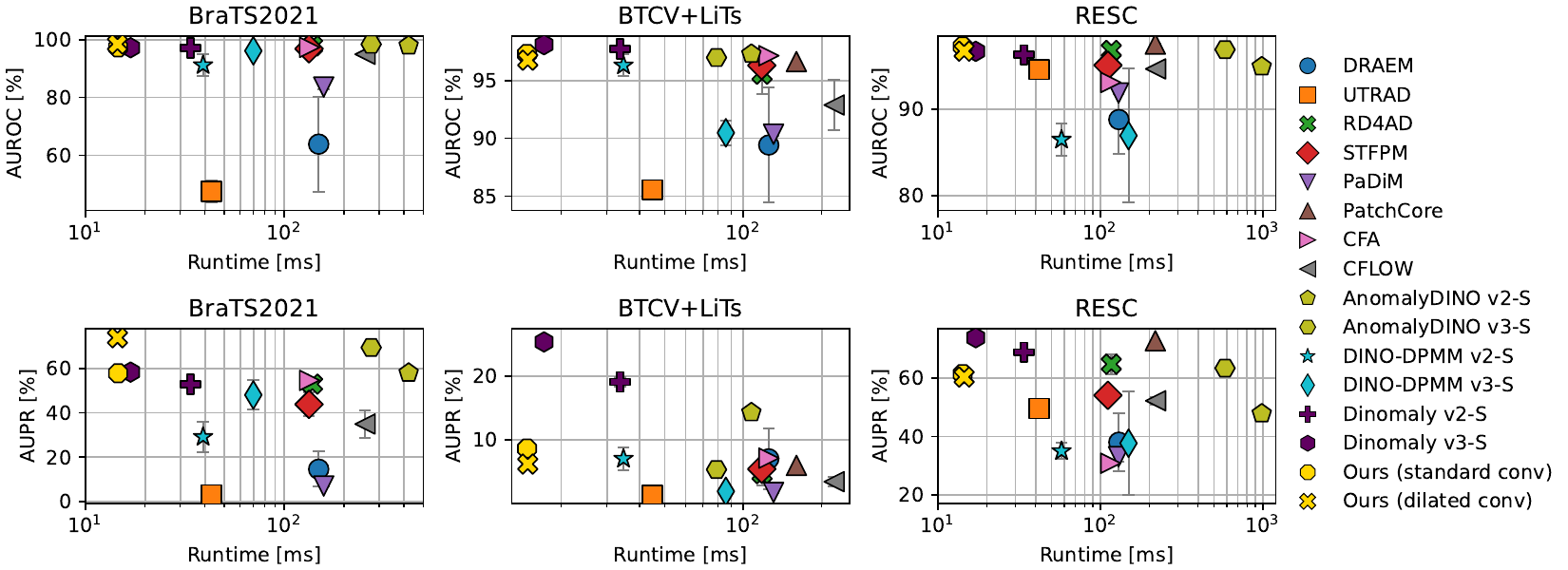}
\caption{
Runtime vs. AUROC and AUPR trade-off across datasets. Methods located in the upper-left region (high detection performance, low runtime) are preferable.
}
\label{fig:results}
\end{figure}

\begin{table}[h!b]
\caption{
Comparison of AUROC and AUPR scores over all datasets and runtime and memory consumption measurements on RESC. AUROC and AUPR scores are averaged over three random seeds and we report mean $\pm$ std. DINO-based methods are highlighted in gray. Results for PatchCore on BraTS2021 are unavailable since this experiment requires more than 80~GB of VRAM. Due to space constraints and little observed variation across datasets, we report runtime and GPU memory measurements only on the RESC dataset.
}
\label{tab:bmad_results}
\centering
\setlength{\tabcolsep}{2pt}
\small
\begin{tabular}{|l|c|rcr|rcr|rcr|c|r|}
\hline
Method                  && \multicolumn{3}{c|}{BraTS2021} & \multicolumn{3}{c|}{BTCV+LiTs} & \multicolumn{3}{c|}{RESC} && \multicolumn{1}{c|}{RESC} \\\hline
DRAEM                   && 63.85 &$\pm$& 16.37 & 89.42 &$\pm$& 4.96 & 88.81 &$\pm$& 3.95 && 130	\\ 
UTRAD                   && 47.47 &$\pm$&  3.78 & 85.54 &$\pm$& 0.36 & 94.62 &$\pm$& 0.11 &&  42 \\
RD4AD                   && 97.85 &$\pm$&  0.05 & 95.54 &$\pm$& 1.73 & 96.84 &$\pm$& 0.23 && 117	\\
STFPM                   && 96.84 &$\pm$&  0.58 & 96.30 &$\pm$& 0.15 & 95.11 &$\pm$& 0.31 && 111	\\
PaDiM                   && 83.59 &$\pm$&  0.80 & 90.35 &$\pm$& 0.17 & 91.88 &$\pm$& 0.07 && 129	\\
PatchCore               && \multicolumn{3}{c|}{N/A} & 96.63 &$\pm$& 0.00 & 97.57 &$\pm$& 0.01 && 218	\\
CFA                     && 97.33 &$\pm$&  0.09 & 97.14 &$\pm$& 0.13 & 93.11 &$\pm$& 0.02 && 116	\\
CFLOW                   && 94.93 &$\pm$&  0.71 & 92.88 &$\pm$& 2.17 & 94.68 &$\pm$& 0.05 && 219	\\
\rowcolor{lightgray!25}
AnomalyDINO (v2-S)      && 97.93 &$\pm$&  0.00 & 97.29 &$\pm$& 0.00 & 94.95 &$\pm$& 0.00 && 984 \\
\rowcolor{lightgray!25}
AnomalyDINO (v3-S)      && 98.38 &$\pm$&  0.00 & 96.98 &$\pm$& 0.00 & 96.93 &$\pm$& 0.00 && 585 \\
\rowcolor{lightgray!25}
DINO-DPMM (v2-S)             && 91.16 &$\pm$&  3.72 & 96.30 &$\pm$& 0.94 & 86.47 &$\pm$& 1.84 &&  58 \\
\rowcolor{lightgray!25}
DINO-DPMM (v3-S)             && 96.12 &$\pm$&  0.55 & 90.48 &$\pm$& 1.05 & 86.94 &$\pm$& 7.76 && 149 \\
\rowcolor{lightgray!25}
Dinomaly (v2-S)         && 97.18 &$\pm$&  0.01 & 97.72 &$\pm$& 0.03 & 96.31	&$\pm$& 0.08 &&  34 \\
\rowcolor{lightgray!25}
Dinomaly (v3-S)         && 97.20 &$\pm$&  0.02 & 98.09 &$\pm$& 0.06 & 96.73	&$\pm$& 0.08 &&  17 \\
\rowcolor{lightgray!25}
Ours (standard conv)    && 97.33 &$\pm$&  0.14 & 97.29 &$\pm$& 0.02 & 97.29 &$\pm$& 0.02 &&  14 \\
\rowcolor{lightgray!25}
Ours (dilated conv)    &\multirow{-13}{*}{\rotatebox{90}{AUROC [\%]}}& 98.44 &$\pm$&  0.04 & 96.75 &$\pm$& 0.04 & 96.75 &$\pm$& 0.04 &\multirow{-13}{*}{\rotatebox{90}{Runtime [ms]}}&  15 \\
\hline\hline
DRAEM                   && 14.61 &$\pm$& 7.94 &  7.03 &$\pm$& 4.79 & 38.05 &$\pm$&  9.94 &&  0.5 \\
UTRAD                   &&  3.11 &$\pm$& 0.25 &  1.36 &$\pm$& 0.04 & 49.57 &$\pm$&  1.20 &&  0.9 \\
RD4AD                   && 52.98 &$\pm$& 1.25 &  4.71 &$\pm$& 1.80 & 64.77 &$\pm$&  3.33 &&  0.4 \\
STFPM                   && 43.86 &$\pm$& 5.28 &  5.44 &$\pm$& 0.29 & 54.10 &$\pm$&  1.02 &&  0.2 \\
PaDiM                   &&  7.02 &$\pm$& 0.32 &  1.76 &$\pm$& 0.03 & 33.09 &$\pm$&  1.73 &&  3.7 \\
PatchCore               && \multicolumn{3}{c|}{N/A} &  6.00 &$\pm$& 0.01 & 72.77 &$\pm$&  0.06 &&  5.0 \\
CFA                     && 54.67 &$\pm$& 0.69 &  7.13 &$\pm$& 0.41 & 30.84 &$\pm$&  0.09 &&  0.3 \\
CFLOW                   && 34.99 &$\pm$& 6.26 &  3.46 &$\pm$& 0.70 & 52.23 &$\pm$&  1.19 &&  0.9 \\
\rowcolor{lightgray!25}
AnomalyDINO (v2-S)      && 58.03 &$\pm$& 0.00 & 14.28 &$\pm$& 0.00 & 47.83 &$\pm$&  0.00 && 38.1 \\
\rowcolor{lightgray!25}
AnomalyDINO (v3-S)      && 69.49 &$\pm$& 0.00 &  5.35 &$\pm$& 0.00 & 63.40 &$\pm$&  0.00 && 11.3 \\
\rowcolor{lightgray!25}
DINO-DPMM (v2-S)             && 29.18 &$\pm$& 6.74 &  7.05 &$\pm$& 1.78 & 34.99 &$\pm$&  2.79 &&  2.0 \\
\rowcolor{lightgray!25}
DINO-DPMM (v3-S)             && 48.09 &$\pm$& 6.62 &  1.94 &$\pm$& 0.25 & 37.70 &$\pm$& 17.83 &&  1.4 \\
\rowcolor{lightgray!25}
Dinomaly (v2-S)         && 52.83 &$\pm$& 0.29 & 19.06 &$\pm$& 0.80 & 68.83 &$\pm$&  0.50 &&  1.3 \\
\rowcolor{lightgray!25}
Dinomaly (v3-S)         && 58.40 &$\pm$& 1.31 & 25.33 &$\pm$& 0.79 & 73.81 &$\pm$&  0.30 &&  0.3 \\
\rowcolor{lightgray!25}
Ours (standard conv)    && 57.91 &$\pm$& 1.93 &  8.64 &$\pm$& 0.24 & 61.38 &$\pm$&  0.61 &&  0.2 \\
\rowcolor{lightgray!25}
Ours (dilated conv)     &\multirow{-14}{*}{\rotatebox{90}{AUPR [\%]}}& 73.88 &$\pm$& 0.41 &  6.21 &$\pm$& 0.18 & 60.36 &$\pm$&  1.05 &\multirow{-14}{*}{\rotatebox{90}{Memory [GB]}}&  0.2 \\
\hline
\end{tabular}
\end{table}

Fig.~\ref{fig:results} illustrates the trade-off between runtime and detection performance across datasets. Our method consistently lies in the favorable region of the plots, achieving competitive AUROC and AUPR while operating at considerably lower runtime compared to all baselines. This highlights the efficiency advantage of AR modeling in embedding space. The detailed quantitative results for BMAD are provided in Tab.~\ref{tab:bmad_results}, some visual results and quantitative results on VisA are provided in Appendix~\ref{sec:app_visual_brats} and~\ref{sec:app_visa}, respectively. 

On the BraTS2021 dataset, the dilated variant of our autoregressive anomaly detection method achieves the best performance in both AUROC and AUPR. In particular, it obtains the highest overall AUPR of 73.88\%, outperforming AnomalyDINO (v3-S) by 4.39 percentage points and surpassing all other baselines. The standard convolution variant also remains competitive by achieving 57.91\% AUPR, which is comparable to strong baselines such as RD4AD (52.98\%) and Dinomaly (v3-S) (58.40\%).

BTCV+LiTs is a challenging dataset on which most methods achieve low AUPR, likely due to an extreme foreground-background imbalance -- where numerous slices contain little to no liver tissue. On this dataset, Dinomaly (v3-S) achieves the best performance with an AUPR of 25.33\%, followed by Dinomaly (v2-S) with 19.06\% and AnomalyDINO (v2-S) with 14.28\%. Our best-performing autoregressive variant on this dataset, using standard convolutions, ranks 4th reaching an AUPR of 8.64\%.

On the RESC dataset, PatchCore achieves the highest AUROC of 97.57\%, while our standard convolution variant obtains a competitive AUROC of 97.29\%, outperforming all other DINO-based methods. In terms of AUPR, Dinomaly (v3-S) achieves the best performance with 73.81\%, followed by PatchCore with 72.77\%. Our autoregressive models reach 61.38\% and 60.36\% AUPR for the standard and dilated variants, respectively, remaining competitive with AnomalyDINO (v3-S) while requiring substantially lower runtime and memory.

One observation from our experiments is that the benefit of dilated convolutions varies across datasets. On BraTS2021 (brain MRI), neighboring patch embeddings are often strongly correlated due to the structured nature of brain anatomy. As a result, the standard AR model may learn a short-range interpolation strategy that generalizes even to anomalous regions, leading to reduced anomaly sensitivity. This interpretation is supported by a retrospective analysis of validation AUROC and AUPR curves: for the standard variant on BraTS, performance reaches a peak in early training and subsequently declines, suggesting overfitting to local interpolation patterns. Incorporating dilated convolutions expands the receptive field and mitigates this effect, improving both AUROC and AUPR. In contrast, for BTCV+LiTs (liver CT) and RESC (retinal OCT), prediction appears to rely more strongly on localized structures, and distant patches provide less informative context. Consequently, enlarging the receptive field does not yield additional benefit and may slightly reduce performance.

\subsection{Ablation}
\paragraph{Comparison with Other Variants} 
Tab.~\ref{tab:ar_bidirectional_multi} compares our AR model with two variants: a bidirectional formulation, where each patch embedding is predicted using both past and future context, and an image space AR model. The bidirectional variant is implemented by masking only the center position in the first convolutional layer and applying standard convolution in subsequent layers. Across datasets, the bidirectional formulation performs comparably to the standard AR model, suggesting that future context offers limited benefit and may occasionally slightly degrade performance. In contrast, performing AR modeling directly in image space leads to a substantial performance drop across all datasets.

\begin{table}[t]
\caption{Comparison of different variants of our method in terms of AUPR [\%].}
\label{tab:ar_bidirectional_multi}
\centering
\small
\setlength{\tabcolsep}{6pt}
\begin{tabular}{|l|c|c|c|}
\hline
Method 
& BraTS2021 
& BTCV+LiTs 
& RESC \\ \hline

Ours (standard conv) 
& 57.91 $\pm$ 1.93 
&  8.64 $\pm$ 0.24
& 61.38 $\pm$ 0.61 \\

Ours (dilated conv)
& 73.88 $\pm$ 0.41 
&  6.21 $\pm$ 0.18
& 60.36 $\pm$ 1.05 \\

Bidirectional (standard conv) 
& 59.61 $\pm$ 0.73 
&  9.66 $\pm$ 0.08
& 61.58 $\pm$ 0.96\\

Bidirectional (dilated conv)
& 72.76 $\pm$ 0.55 
&  8.35 $\pm$ 0.04
& 65.56 $\pm$ 0.26 \\

Image space AR
& 11.03 $\pm$ 0.15 
& 4.14 $\pm$ 0.06 
& 15.10 $\pm$ 0.05 \\ \hline

\end{tabular}
\end{table}

\paragraph{Effect of Using Larger DINOv3 Models}
We evaluate the impact of backbone scale by replacing DINOv3-S with DINOv3-7B in our method on BraTS2021. The larger backbone yields only marginal differences (AUROC: $98.44 \rightarrow 98.38$, AUPR: $73.88 \rightarrow 74.56$) while substantially increasing runtime. This suggests that scaling the DINO backbone alone provides limited benefit for anomaly detection in our autoregressive framework.

\section{Conclusion}
We presented a simple and efficient framework for UAD by performing spatial AR modeling directly on DINOv3 patch embeddings. By explicitly capturing conditional dependencies on the 2D embedding grid with a lightweight AR CNN, our approach avoids memory-bank storage and costly nearest-neighbor retrieval, enabling fast and memory-efficient inference with a single forward pass. Experiments on the BMAD benchmark~\cite{bao2024bmad} demonstrate competitive UAD performance while substantially reducing runtime and GPU memory usage on brain MR and retinal OCT images. A similar behavior can be observed for the VisA dataset. The UAD performance of the proposed method only lags behind state-of-the-art methods for the liver CT benchmark in BMAD, which is a particularly challenging benchmark where none of the evaluated methods reach satisfactory UAD performances.

Currently, the proposed method processes medical images on a slice-by-slice basis. Incorporating volumetric (3D) context is expected to significantly enhance accuracy, particularly for the BTCV+LiTs datasets.

Another limitation is the setup in BMAD, where slices of a volume not containing any tumor or retinal edema annotations are treated as normal, while slices with annotations are treated as abnormal. This leads to the central part of the brain being underrepresented by the healthy slices and overrepresented by abnormal slices, which biases the UAD methods towards better results. On top, regions with deformed tissues that are not part of the tumor itself are treated as normal, even though these deformation are not common in healthy brains. This could be mitigated by using separate datasets for healthy distribution modeling as proposed in~\cite{frotscher2025deep}. In addition, this will allow to better asses UAD performance on a patient level, which is crucial for clinical applicability.

Furthermore, selecting the optimal configuration between dilated and standard convolutions as well as the causal and the bidirectional factorization across diverse datasets remains challenging and a comprehensive comparative analysis is left for future research.

\begin{credits}
\subsubsection{\ackname} 
We acknowledge The LOOP Zurich - Medical Research Center, Zurich, Switzerland, Georg and Berta Schwyzer-Winiker Foundation, and Innovationspool University Zurich Hospital for the financial support for this project.

\subsubsection{\discintname}
The authors have no competing interests to declare that are relevant to the content of this article.
\end{credits}

%
%
\clearpage
\appendix
\section*{Appendix}
\input{appendix_arxiv}

\clearpage
\newpage
\bibliographystyle{splncs04}
\bibliography{refs}

\end{document}

%% file: preamble.tex
\usepackage[T1]{fontenc}
\usepackage{graphicx,verbatim}
\usepackage{amsfonts}
\usepackage{amsmath}
\usepackage{subcaption} 
\usepackage{rotating} 
\usepackage{xcolor} 
\usepackage[table]{xcolor}
\usepackage[hidelinks]{hyperref}
\usepackage{tabularx} 
\usepackage{multirow} 
\usepackage{array}
\newcommand{\PreserveBackslash}[1]{\let\temp=\\#1\let\\=\temp}
\newcolumntype{C}[1]{>{\PreserveBackslash\centering}p{#1}}
\newcolumntype{R}[1]{>{\PreserveBackslash\raggedleft}p{#1}}
\newcolumntype{L}[1]{>{\PreserveBackslash\raggedright}p{#1}}
\usepackage{orcidlink}
\usepackage[misc,geometry]{ifsym}
\author{Ertunc Erdil\orcidlink{0000-0001-6235-4574}\inst{1}\textsuperscript{*} \and
Nico Schulthess\orcidlink{0000-0002-3421-5638}\inst{1}\textsuperscript{*} \and
Guney I. Tombak\orcidlink{0000-0001-7852-0696}\inst{1} \and
Ender Konukoglu\orcidlink{0000-0002-2542-3611}\inst{1,2}}
\authorrunning{E. Erdil and N. Schulthess et al.}
\institute{
Biomedical Image Computing Group, ETH Zurich, Zurich, Switzerland \and
The LOOP Zurich - Medical Research Center, Zurich, Switzerland\\
\textsuperscript{*} Authors with equal contribution\\
\email{erdile@ethz.ch}
}

%% file: appendix_arxiv.tex
\section{Visual Results on BraTS}
\label{sec:app_visual_brats}

\begin{figure}[h!]
\centering
\begin{subfigure}[c]{0.02\textwidth}
	\begin{minipage}[c][5\textwidth][c]{\textwidth}
	\begin{turn}{90}\small
		Image
	\end{turn}
	\end{minipage}
	\begin{minipage}[c][5\textwidth][c]{\textwidth}
	\begin{turn}{90}\small
		Labels
	\end{turn}
	\end{minipage}
	\begin{minipage}[c][5\textwidth][c]{\textwidth}
	\begin{turn}{90}\small
		Ours
	\end{turn}
	\end{minipage}
	\begin{minipage}[c][5\textwidth][c]{\textwidth}
	\hspace{-\baselineskip}\begin{turn}{90}\small
		\shortstack{Anomaly-\\DINO}
	\end{turn}
	\end{minipage}
	\begin{minipage}[c][5\textwidth][c]{\textwidth}
	\begin{turn}{90}\small
		Dinomaly
	\end{turn}
	\end{minipage}
\end{subfigure}
\begin{subfigure}[c]{0.1\textwidth}
	\includegraphics[angle=270, origin=c, width=\textwidth]{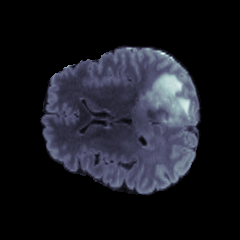}
    \includegraphics[angle=270, origin=c, width=\textwidth]{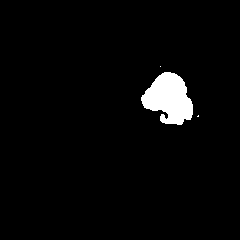}
	\includegraphics[angle=270, origin=c, width=\textwidth]{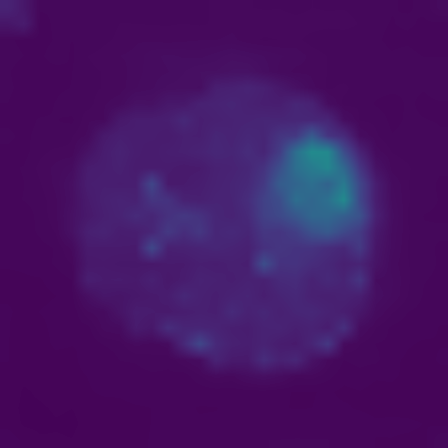}
	\includegraphics[angle=  0, origin=c, width=\textwidth]{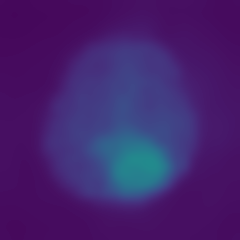}
	\includegraphics[angle=270, origin=c, width=\textwidth]{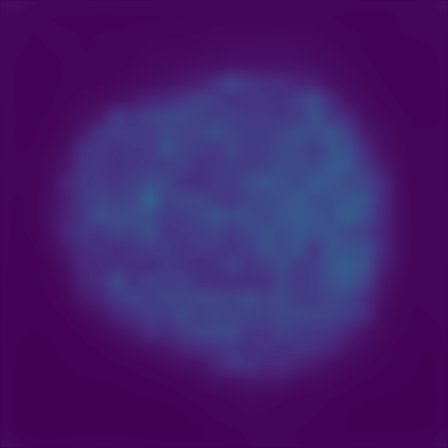}
\end{subfigure}
\begin{subfigure}[c]{0.1\textwidth}
	\includegraphics[angle=270, origin=c, width=\textwidth]{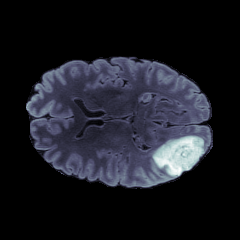}
	\includegraphics[angle=270, origin=c, width=\textwidth]{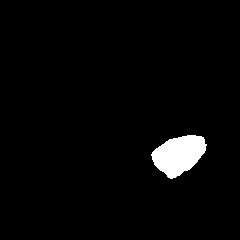}
	\includegraphics[angle=270, origin=c, width=\textwidth]{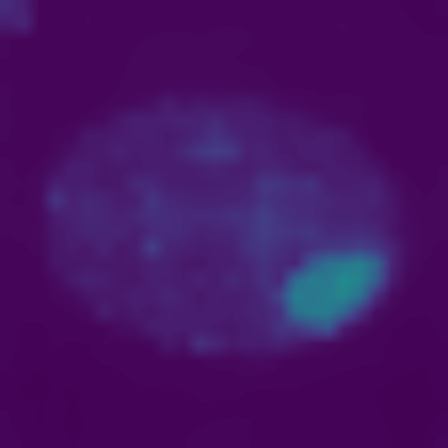}
	\includegraphics[angle=  0, origin=c, width=\textwidth]{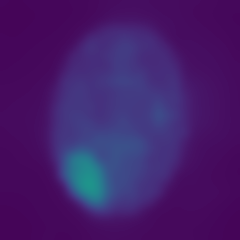}
	\includegraphics[angle=270, origin=c, width=\textwidth]{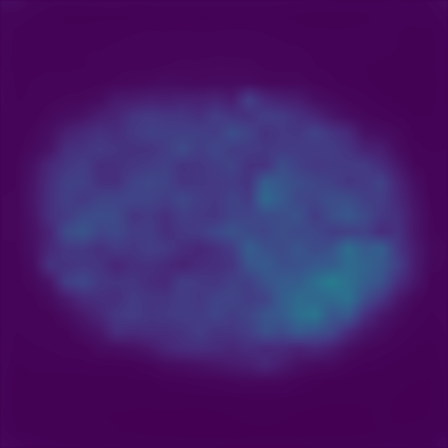}
\end{subfigure}
\begin{subfigure}[c]{0.1\textwidth}
	\includegraphics[angle=270, origin=c, width=\textwidth]{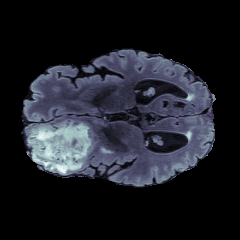}
	\includegraphics[angle=270, origin=c, width=\textwidth]{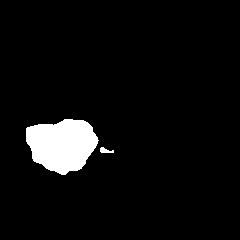}
	\includegraphics[angle=270, origin=c, width=\textwidth]{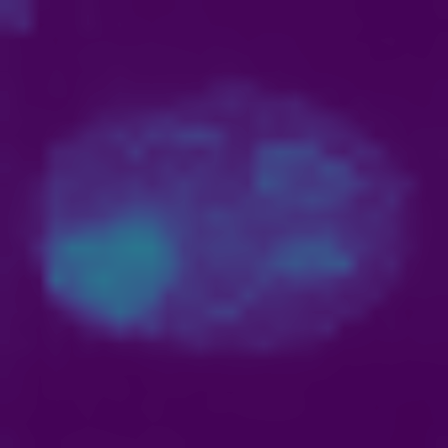}
	\includegraphics[angle=  0, origin=c, width=\textwidth]{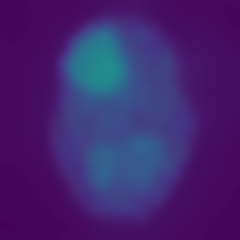}
	\includegraphics[angle=270, origin=c, width=\textwidth]{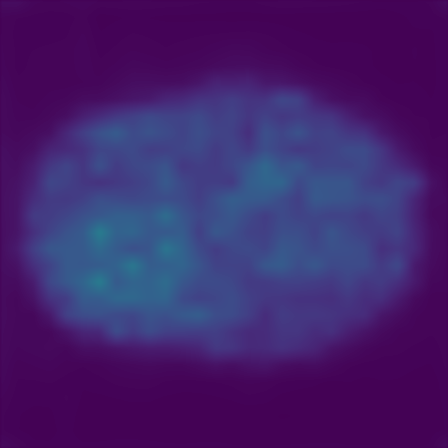}
\end{subfigure}
\begin{subfigure}[c]{0.1\textwidth}
	\includegraphics[angle=270, origin=c, width=\textwidth]{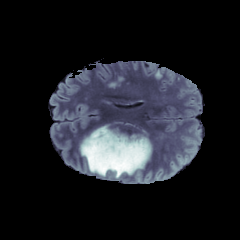}
	\includegraphics[angle=270, origin=c, width=\textwidth]{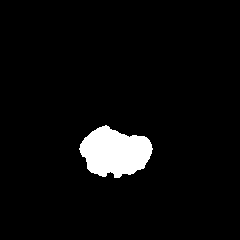}
	\includegraphics[angle=270, origin=c, width=\textwidth]{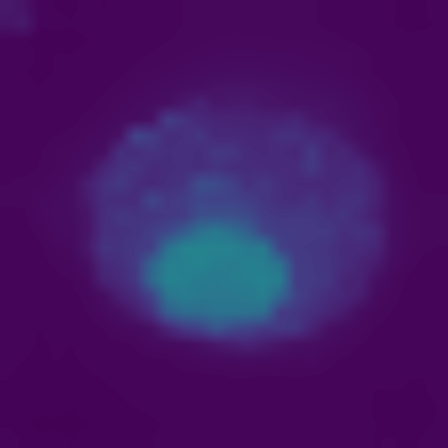}
	\includegraphics[angle=  0, origin=c, width=\textwidth]{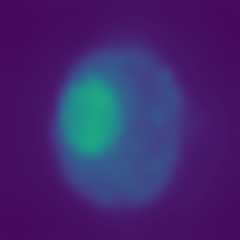}
	\includegraphics[angle=270, origin=c, width=\textwidth]{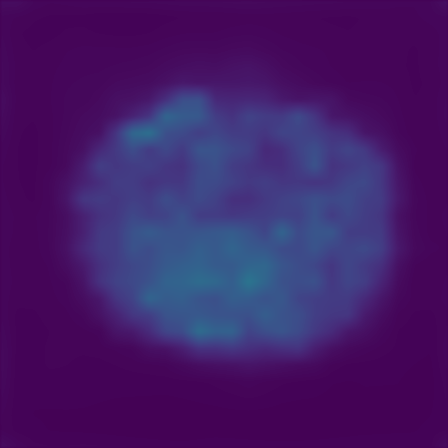}
\end{subfigure}
\begin{subfigure}[c]{0.1\textwidth}
	\includegraphics[angle=270, origin=c, width=\textwidth]{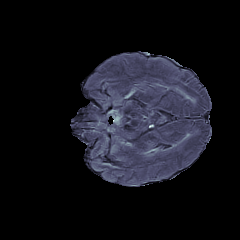}
	\includegraphics[angle=270, origin=c, width=\textwidth]{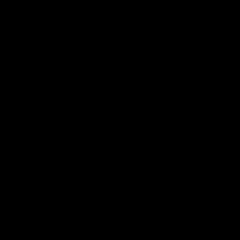}
	\includegraphics[angle=270, origin=c, width=\textwidth]{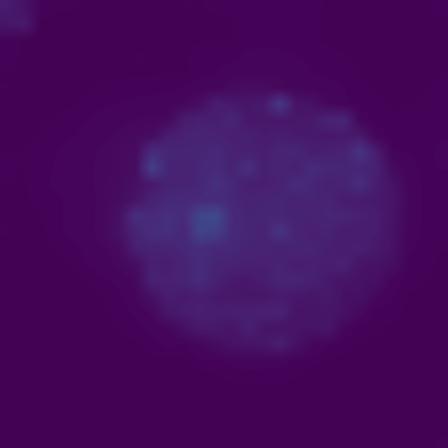}
	\includegraphics[angle=  0, origin=c, width=\textwidth]{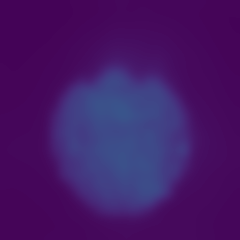}
	\includegraphics[angle=270, origin=c, width=\textwidth]{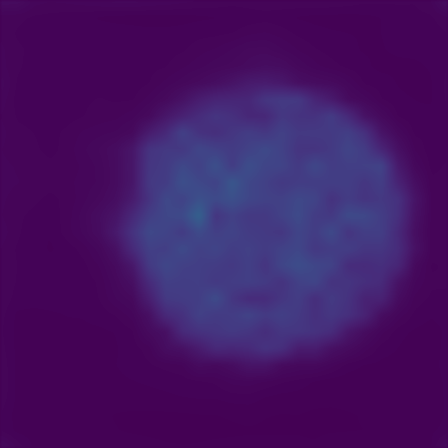}
\end{subfigure}
\begin{subfigure}[c]{0.1\textwidth}
	\includegraphics[angle=270, origin=c, width=\textwidth]{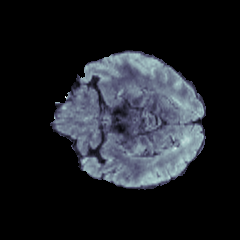}
	\includegraphics[angle=270, origin=c, width=\textwidth]{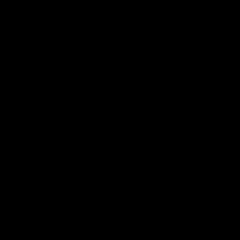}
	\includegraphics[angle=270, origin=c, width=\textwidth]{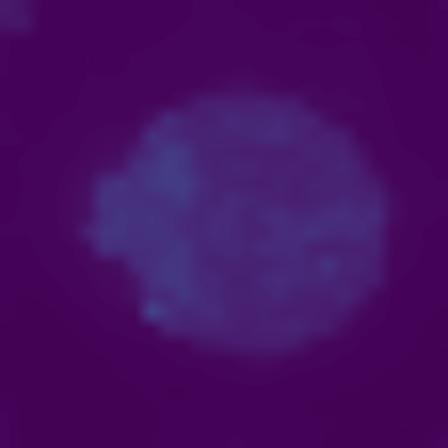}
	\includegraphics[angle=  0, origin=c, width=\textwidth]{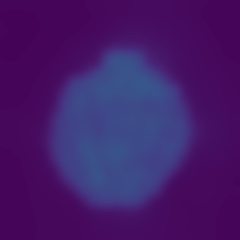}
	\includegraphics[angle=270, origin=c, width=\textwidth]{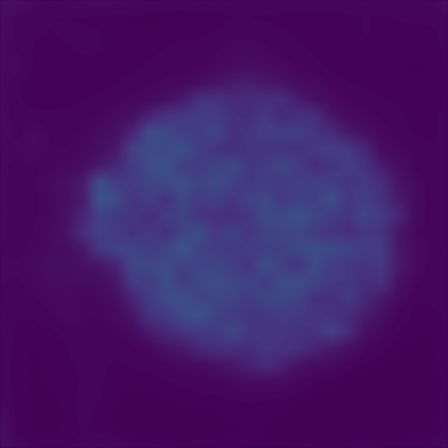}
\end{subfigure}
\begin{subfigure}[c]{0.1\textwidth}
	\includegraphics[angle=270, origin=c, width=\textwidth]{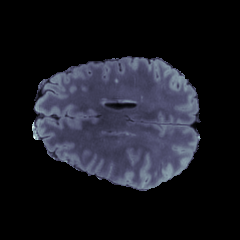}
	\includegraphics[angle=270, origin=c, width=\textwidth]{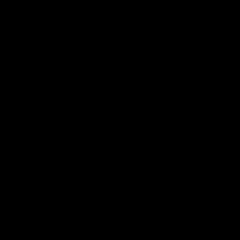}
	\includegraphics[angle=270, origin=c, width=\textwidth]{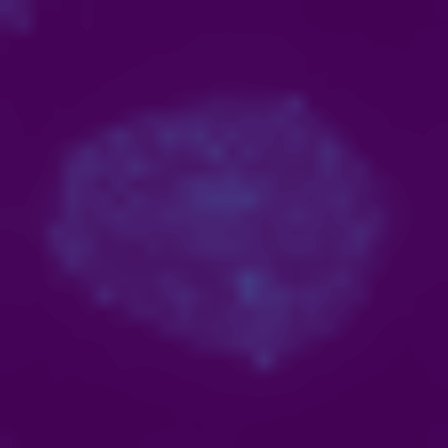}
	\includegraphics[angle=  0, origin=c, width=\textwidth]{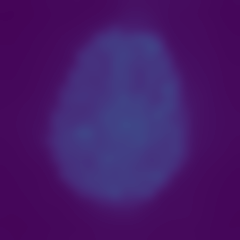}
	\includegraphics[angle=270, origin=c, width=\textwidth]{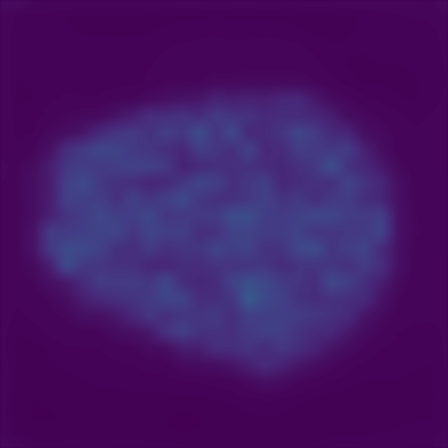}
\end{subfigure}
\begin{subfigure}[c]{0.1\textwidth}
	\includegraphics[angle=270, origin=c, width=\textwidth]{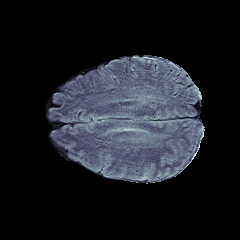}
	\includegraphics[angle=270, origin=c, width=\textwidth]{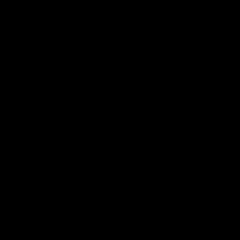}
	\includegraphics[angle=270, origin=c, width=\textwidth]{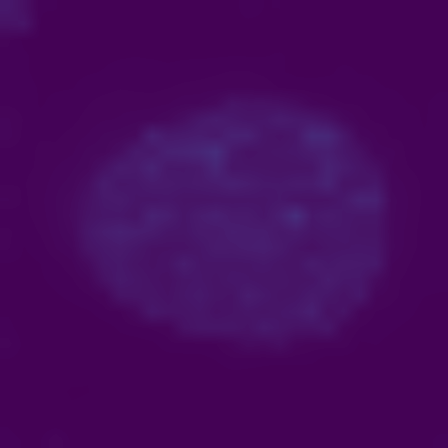}
	\includegraphics[angle=  0, origin=c, width=\textwidth]{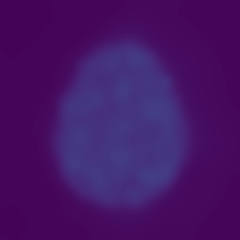}
	\includegraphics[angle=270, origin=c, width=\textwidth]{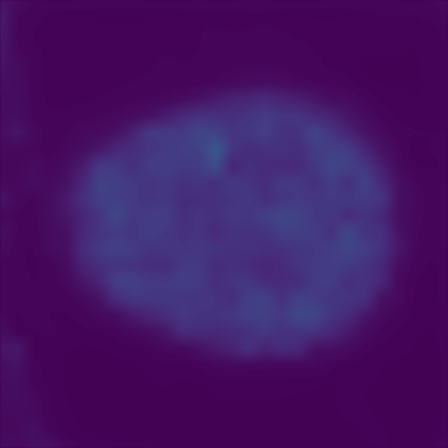}
\end{subfigure}
\caption{Visual results for our method, AnomalyDINO and Dinomaly.}
\label{fig:visual_results}
\end{figure}

\vspace{-1.0cm}
\section{Results on VisA}
\label{sec:app_visa}
In addition to the medical datasets, we evaluated the DINO-based methods on the industrial VisA dataset. Since there is no validation set available, no model selection was performed and we evaluated the model after the last epoch. Otherwise, the evaluation setting is the same as for the BMAD datasets.
AUROC and AUPR scores averaged over three seeds and all 12 subsets are provided in Tab.~\ref{tab:visa_results}.
\begin{table}[b!h]
\caption{
AUROC and AUPR results on VisA for DINO-based methods, averaged across all object categories. For each method, scores are computed over three random seeds and reported as mean $\pm$ standard deviation.
}
\label{tab:visa_results}
\centering
\setlength{\tabcolsep}{2pt}
\small
\begin{tabular}{|l|rcr|rcr|}
\hline
Method                  & \multicolumn{3}{c|}{AUROC} & \multicolumn{3}{c|}{AUPR} \\\hline
AnomalyDINO (v2-S)      & 98.88 &$\pm$& 0.00 & 41.62 &$\pm$& 0.00 \\
AnomalyDINO (v3-S)      & 99.01 &$\pm$& 0.00 & 44.23 &$\pm$& 0.00 \\
DINO-DPMM (v2-S)        & 30.73 &$\pm$& 2.76 &  0.48 &$\pm$& 0.09 \\
DINO-DPMM (v3-S)        & 27.34 &$\pm$& 0.66 &  0.39 &$\pm$& 0.00 \\
Dinomaly (v2-S)         & 96.70 &$\pm$& 0.06 & 29.04 &$\pm$& 0.52 \\
Dinomaly (v3-S)         & 99.15 &$\pm$& 0.02 & 44.49 &$\pm$& 0.39 \\
Ours (standard conv)    & 98.76 &$\pm$& 0.01 & 43.82 &$\pm$& 0.49 \\ \hline
\end{tabular}
\end{table}

%% file: refs.bib
@article{chen2018unsupervised,
  title={Unsupervised detection of lesions in brain MRI using constrained adversarial auto-encoders},
  author={Chen, Xiaoran and Konukoglu, Ender},
  journal={arXiv preprint arXiv:1806.04972},
  year={2018}
}

@inproceedings{zimmerer2019unsupervised,
  title={Unsupervised anomaly localization using variational auto-encoders},
  author={Zimmerer, David and Isensee, Fabian and Petersen, Jens and Kohl, Simon and Maier-Hein, Klaus},
  booktitle={International conference on medical image computing and computer-assisted intervention},
  pages={289--297},
  year={2019},
  organization={Springer}
}

@article{pinaya2022unsupervised,
    title={Unsupervised brain imaging 3D anomaly detection and segmentation with transformers},
    author={Pinaya, Walter HL and Tudosiu, Petru-Daniel and Gray, Robert and Rees, Geraint and Nachev, Parashkev and Ourselin, Sebastien and Cardoso, M Jorge},
    journal={Medical Image Analysis},
    volume={79},
    pages={102475},
    year={2022},
    publisher={Elsevier}
}

@inproceedings{marimont2021anomaly,
    title={Anomaly detection through latent space restoration using vector quantized variational autoencoders},
    author={Marimont, Sergio Naval and Tarroni, Giacomo},
    booktitle={2021 IEEE 18th International Symposium on Biomedical Imaging (ISBI)},
    pages={1764--1767},
    year={2021},
    organization={IEEE}
}

@article{schlegl2019f,
  title={f-AnoGAN: Fast unsupervised anomaly detection with generative adversarial networks},
  author={Schlegl, Thomas and Seeb{\"o}ck, Philipp and Waldstein, Sebastian M and Langs, Georg and Schmidt-Erfurth, Ursula},
  journal={Medical image analysis},
  volume={54},
  pages={30--44},
  year={2019},
  publisher={Elsevier}
}

@inproceedings{wyatt2022anoddpm,
  title={Anoddpm: Anomaly detection with denoising diffusion probabilistic models using simplex noise},
  author={Wyatt, Julian and Leach, Adam and Schmon, Sebastian M and Willcocks, Chris G},
  booktitle={Proceedings of the IEEE/CVF conference on computer vision and pattern recognition},
  pages={650--656},
  year={2022}
}

@inproceedings{zavrtanik2021draem,
  title={Draem-a discriminatively trained reconstruction embedding for surface anomaly detection},
  author={Zavrtanik, Vitjan and Kristan, Matej and Sko{\v{c}}aj, Danijel},
  booktitle={Proceedings of the IEEE/CVF international conference on computer vision},
  pages={8330--8339},
  year={2021}
}

@article{chen2022utrad,
  title={UTRAD: Anomaly detection and localization with U-transformer},
  author={Chen, Liyang and You, Zhiyuan and Zhang, Nian and Xi, Juntong and Le, Xinyi},
  journal={Neural Networks},
  volume={147},
  pages={53--62},
  year={2022},
  publisher={Elsevier}
}

@inproceedings{gudovskiy2022cflow,
  title={Cflow-ad: Real-time unsupervised anomaly detection with localization via conditional normalizing flows},
  author={Gudovskiy, Denis and Ishizaka, Shun and Kozuka, Kazuki},
  booktitle={Proceedings of the IEEE/CVF winter conference on applications of computer vision},
  pages={98--107},
  year={2022}
}

@inproceedings{deng2022anomaly,
  title={Anomaly detection via reverse distillation from one-class embedding},
  author={Deng, Hanqiu and Li, Xingyu},
  booktitle={Proceedings of the IEEE/CVF conference on computer vision and pattern recognition},
  pages={9737--9746},
  year={2022}
}

@article{wang2021student,
  title={Student-teacher feature pyramid matching for anomaly detection},
  author={Wang, Guodong and Han, Shumin and Ding, Errui and Huang, Di},
  journal={arXiv preprint arXiv:2103.04257},
  year={2021}
}

@inproceedings{defard2021padim,
  title={Padim: a patch distribution modeling framework for anomaly detection and localization},
  author={Defard, Thomas and Setkov, Aleksandr and Loesch, Angelique and Audigier, Romaric},
  booktitle={International conference on pattern recognition},
  pages={475--489},
  year={2021},
  organization={Springer}
}

@article{cohen2020sub,
  title={Sub-image anomaly detection with deep pyramid correspondences},
  author={Cohen, Niv and Hoshen, Yedid},
  journal={arXiv preprint arXiv:2005.02357},
  year={2020}
}

@inproceedings{roth2022towards,
  title={Towards total recall in industrial anomaly detection},
  author={Roth, Karsten and Pemula, Latha and Zepeda, Joaquin and Sch{\"o}lkopf, Bernhard and Brox, Thomas and Gehler, Peter},
  booktitle={Proceedings of the IEEE/CVF conference on computer vision and pattern recognition},
  pages={14318--14328},
  year={2022}
}

@article{huang2024prototype,
  title={A prototype-based neural network for image anomaly detection and localization},
  author={Huang, Chao and Kang, Zhao and Wu, Hong},
  journal={Neural Processing Letters},
  volume={56},
  number={3},
  pages={169},
  year={2024},
  publisher={Springer}
}

@article{lee2022cfa,
  title={Cfa: Coupled-hypersphere-based feature adaptation for target-oriented anomaly localization},
  author={Lee, Sungwook and Lee, Seunghyun and Song, Byung Cheol},
  journal={IEEE Access},
  volume={10},
  pages={78446--78454},
  year={2022},
  publisher={IEEE}
}

@inproceedings{damm2025anomalydino,
  title={Anomalydino: Boosting patch-based few-shot anomaly detection with dinov2},
  author={Damm, Simon and Laszkiewicz, Mike and Lederer, Johannes and Fischer, Asja},
  booktitle={2025 IEEE/CVF Winter Conference on Applications of Computer Vision (WACV)},
  pages={1319--1329},
  year={2025},
  organization={IEEE}
}

@inproceedings{schulthess2025anomaly,
  title={Anomaly Detection by Clustering DINO Embeddings Using a Dirichlet Process Mixture},
  author={Schulthess, Nico and Konukoglu, Ender},
  booktitle={International Conference on Medical Image Computing and Computer-Assisted Intervention},
  pages={46--56},
  year={2025},
  organization={Springer}
}

@inproceedings{guo2025dinomaly,
  title={Dinomaly: The less is more philosophy in multi-class unsupervised anomaly detection},
  author={Guo, Jia and Lu, Shuai and Zhang, Weihang and Chen, Fang and Li, Huiqi and Liao, Hongen},
  booktitle={Proceedings of the Computer Vision and Pattern Recognition Conference},
  pages={20405--20415},
  year={2025}
}

@article{oquab2023dinov2,
  title={Dinov2: Learning robust visual features without supervision},
  author={Oquab, Maxime and Darcet, Timoth{\'e}e and Moutakanni, Th{\'e}o and Vo, Huy and Szafraniec, Marc and Khalidov, Vasil and Fernandez, Pierre and Haziza, Daniel and Massa, Francisco and El-Nouby, Alaaeldin and others},
  journal={arXiv preprint arXiv:2304.07193},
  year={2023}
}

@article{simeoni2025dinov3,
  title={Dinov3},
  author={Sim{\'e}oni, Oriane and Vo, Huy V and Seitzer, Maximilian and Baldassarre, Federico and Oquab, Maxime and Jose, Cijo and Khalidov, Vasil and Szafraniec, Marc and Yi, Seungeun and Ramamonjisoa, Micha{\"e}l and others},
  journal={arXiv preprint arXiv:2508.10104},
  year={2025}
}

@inproceedings{bao2024bmad,
  title={Bmad: Benchmarks for medical anomaly detection},
  author={Bao, Jinan and Sun, Hanshi and Deng, Hanqiu and He, Yinsheng and Zhang, Zhaoxiang and Li, Xingyu},
  booktitle={Proceedings of the IEEE/CVF conference on computer vision and pattern recognition},
  pages={4042--4053},
  year={2024}
}

@article{baid2021rsna,
  title={The rsna-asnr-miccai brats 2021 benchmark on brain tumor segmentation and radiogenomic classification},
  author={Baid, Ujjwal and Ghodasara, Satyam and Mohan, Suyash and Bilello, Michel and Calabrese, Evan and Colak, Errol and Farahani, Keyvan and Kalpathy-Cramer, Jayashree and Kitamura, Felipe C and Pati, Sarthak and others},
  journal={arXiv preprint arXiv:2107.02314},
  year={2021}
}

@misc{landman2015miccai,
  title={miccai multi-atlas labeling beyond the cranial vault--workshop and challenge. Accessed Dec. 2020},
  author={Landman, BA and Xu, Z and Igelsias, JE and Styner, M and Langerak, TR and Klein, A},
  year={2015}
}

@article{bilic2023liver,
  title={The liver tumor segmentation benchmark (lits)},
  author={Bilic, Patrick and Christ, Patrick and Li, Hongwei Bran and Vorontsov, Eugene and Ben-Cohen, Avi and Kaissis, Georgios and Szeskin, Adi and Jacobs, Colin and Mamani, Gabriel Efrain Humpire and Chartrand, Gabriel and others},
  journal={Medical image analysis},
  volume={84},
  pages={102680},
  year={2023},
  publisher={Elsevier}
}

@article{hu2019automated,
  title={Automated segmentation of macular edema in OCT using deep neural networks},
  author={Hu, Junjie and Chen, Yuanyuan and Yi, Zhang},
  journal={Medical image analysis},
  volume={55},
  pages={216--227},
  year={2019},
  publisher={Elsevier}
}

@inproceedings{zou2022spot,
  title={Spot-the-difference self-supervised pre-training for anomaly detection and segmentation},
  author={Zou, Yang and Jeong, Jongheon and Pemula, Latha and Zhang, Dongqing and Dabeer, Onkar},
  booktitle={European conference on computer vision},
  pages={392--408},
  year={2022},
  organization={Springer}
}

@article{frotscher2025deep,
    title={Deep Unsupervised Anomaly Detection in Brain Imaging: Large-Scale Benchmarking and Bias Analysis}, 
    author={Alexander Frotscher and Christian F. Baumgartner and Thomas Wolfers},
    year={2025},
    journal={arXiv preprint arXiv:2512.01534},
}

@inproceedings{van2016pixel,
  title={Pixel recurrent neural networks},
  author={Van Den Oord, A{\"a}ron and Kalchbrenner, Nal and Kavukcuoglu, Koray},
  booktitle={International conference on machine learning},
  pages={1747--1756},
  year={2016},
  organization={PMLR}
}

@article{ren2019likelihood,
    title={Likelihood ratios for out-of-distribution detection},
    author={Ren, Jie and Liu, Peter J and Fertig, Emily and Snoek, Jasper and Poplin, Ryan and Depristo, Mark and Dillon, Joshua and Lakshminarayanan, Balaji},
    journal={Advances in neural information processing systems},
    volume={32},
    year={2019}
}

@article{serra2019input,
    title={Input complexity and out-of-distribution detection with likelihood-based generative models},
    author={Serr{\`a}, Joan and {\'A}lvarez, David and G{\'o}mez, Vicen{\c{c}} and Slizovskaia, Olga and N{\'u}{\~n}ez, Jos{\'e} F and Luque, Jordi},
    journal={arXiv preprint arXiv:1909.11480},
    year={2019}
}

@inproceedings{nalisnick2018do,
    title={Do Deep Generative Models Know What They Don't Know?},
    author={Eric Nalisnick and Akihiro Matsukawa and Yee Whye Teh and Dilan Gorur and Balaji Lakshminarayanan},
    booktitle={International Conference on Learning Representations},
    year={2019},
    url={https://openreview.net/forum?id=H1xwNhCcYm},
}

@article{schirrmeister2020understanding,
    title={Understanding anomaly detection with deep invertible networks through hierarchies of distributions and features},
    author={Schirrmeister, Robin and Zhou, Yuxuan and Ball, Tonio and Zhang, Dan},
    journal={Advances in Neural Information Processing Systems},
    volume={33},
    pages={21038--21049},
    year={2020}
}

@article{umapathi2022shaken,
    title={Shaken, and Stirred: Long-Range Dependencies Enable Robust Outlier Detection with PixelCNN++},
    author={Umapathi, Barath Mohan and Chauhan, Kushal and Shenoy, Pradeep and Sridharan, Devarajan},
    journal={arXiv preprint arXiv:2208.13579},
    year={2022}
}

@article{yu2015multi,
  title={Multi-scale context aggregation by dilated convolutions},
  author={Yu, Fisher and Koltun, Vladlen},
  journal={arXiv preprint arXiv:1511.07122},
  year={2015}
}

@article{loshchilov2017decoupled,
  title={Decoupled weight decay regularization},
  author={Loshchilov, Ilya and Hutter, Frank},
  journal={arXiv preprint arXiv:1711.05101},
  year={2017}
}

@article{xiong2024autoregressive,
  title={Autoregressive models in vision: A survey},
  author={Xiong, Jing and Liu, Gongye and Huang, Lun and Wu, Chengyue and Wu, Taiqiang and Mu, Yao and Yao, Yuan and Shen, Hui and Wan, Zhongwei and Huang, Jinfa and others},
  journal={arXiv preprint arXiv:2411.05902},
  year={2024}
}

@incollection{tomczak2024autoregressive,
  title={Autoregressive Models},
  author={Tomczak, Jakub M},
  booktitle={Deep Generative Modeling},
  pages={37--62},
  year={2024},
  publisher={Springer}
}

@article{sun2024autoregressive,
  title={Autoregressive model beats diffusion: Llama for scalable image generation},
  author={Sun, Peize and Jiang, Yi and Chen, Shoufa and Zhang, Shilong and Peng, Bingyue and Luo, Ping and Yuan, Zehuan},
  journal={arXiv preprint arXiv:2406.06525},
  year={2024}
}

@inproceedings{Zagoruyko2016Wide,
  title = {Wide Residual Networks},
  booktitle = {Proceedings of the British Machine Vision Conference ({{BMVC}})},
  author = {Zagoruyko, Sergey and Komodakis, Nikos},
  year = {2016},
  pages = {87.1-87.12},
  articleno = {87}
}
